\documentclass[runningheads]{llncs}
\usepackage[T1]{fontenc}
\usepackage{graphicx}
\usepackage{booktabs}
\usepackage[misc]{ifsym}
\usepackage{adjustbox}
\usepackage{multirow}
\usepackage{amsmath}
\usepackage{dblfloatfix}
\usepackage[misc]{ifsym}
\usepackage[sectionbib]{chapterbib}
\usepackage{appendix}

\usepackage{xurl}
\newcommand{\corr}{(\Letter)}
\usepackage{mwe}

\begin{document}

\title{MT-HCCAR: Multi-Task Deep Learning with Hierarchical Classification and Attention-based Regression for Cloud Property Retrieval}

\titlerunning{Multi-Task Deep Learning for Cloud Property Retrieval}

\author{Xingyan Li\inst{1}\orcidID{0009-0001-2598-2296} \and
Andrew M. Sayer\inst{2,3}\orcidID{0000-0001-9149-1789} \and
Ian T. Carroll\inst{2,3} \and 
Xin Huang\inst{4} \and
Jianwu Wang\inst{1}\orcidID{0000-0002-9933-1170}\corr}

\authorrunning{X. Li et al.}

\institute{Department of Information Systems, University of Maryland, Baltimore County, Baltimore, MD, USA
\email{\{xingyanli,jianwuw\}@umbc.edu}
\and
NASA Goddard Space Flight Center, Greenbelt, MD, USA 
\and
Goddard Earth Sciences Technology and Research (GESTAR) II, UMBC, Baltimore, MD, USA
\and
Dept of Computer \& Information Sciences, Towson University, Towson, MD, USA
}

\maketitle              

\begin{abstract}
In Earth science, accurate retrieval of cloud properties including cloud masking, cloud phase classification, and cloud optical thickness (COT) prediction is essential in atmospheric and environmental studies. Conventional methods rely on distinct models for each sensor due to their unique spectral characteristics. Recently, machine/deep learning has been embraced to extract features from satellite datasets, yet existing approaches lack architectures capturing hierarchical relationships among tasks. Additionally, given the spectral diversity among sensors, developing models with robust generalization capabilities remains challenging for related research. There is also a notable absence of methods evaluated across different satellite sensors. In response, we propose MT-HCCAR, an end-to-end deep learning model employing multi-task learning. MT-HCCAR simultaneously handles cloud masking, cloud phase retrieval (classification tasks), and COT prediction (a regression task). It integrates a hierarchical classification network (HC) and a classification-assisted attention-based regression network (CAR), enhancing precision and robustness in cloud labeling and COT prediction. Experimental evaluations, including comparisons with baseline methods and ablation studies, demonstrate that MT-HCCAR achieves optimal performance across various evaluation metrics and satellite datasets. 

\keywords{Cloud property retrieval  \and Remote sensing \and Deep learning.}
\end{abstract}

\section{Introduction}
Clouds are integral components of the Earth system, wielding substantial influence over the planet's energy dynamics, climate regulation, and the hydrological cycle~\cite{voigt2021clouds}. Satellites have long been an indispensable tool to help us understand our environment. A prominent category of these instruments, commonly referred to as imagers, passively collect measurements of the Earth across various combinations of ultraviolet (UV), visible (VIS), near-infrared (NIR), shortwave infrared (SWIR), and thermal infrared (TIR) wavelength ranges.  These measurements of reflected solar and/or emitted thermal radiation in different spectral bands undergo routine processing by algorithms to convert them into geophysical parameters of interest (atmospheric and surface characteristics) in a process known as ``retrieval''. This is typically done pixel-by-pixel for each satellite image.  For clouds, several key properties are targeted, including a cloud mask (which distinguishes cloud-covered pixels from cloud-free ones), thermodynamic phase (indicating whether the cloud comprises liquid water droplets or ice crystals), and cloud optical thickness (COT), which is a measure of both the amount of light scattered by a cloud and the quantity of liquid or ice within it. The routine retrieval of geophysical parameters is essential for advancing our understanding of Earth's climate~\cite{Hollmann2013}.

One practical motivation for our work is NASA’s Plankton, Aerosol, Cloud, ocean Ecosystem (PACE) mission~\cite{Werdell2019}, which launched in February 2024. Existing NASA cloud masking algorithms are not directly applicable to PACE’s main sensor, called OCI~\cite{OCI-site}, due to different spectral bands compared to algorithms developed for existing sensors. OCI has similarities with some previous spaceborne imagers such as MODIS~\cite{MODIS-site}, VIIRS~\cite{VIIRS-site}, and ABI~\cite{ABI-site} sensor types. However, two key differences are that 1) OCI has continuous hyperspectral coverage in the UV-NIR, plus discrete SWIR bands, while the others are only multi-spectral (up to a dozen or so discrete relevant bands); and 2) OCI lacks TIR bands. Some of the most commonly used cloud masking tests for those sensors are therefore not applicable, and adapting a subset of those tests would miss out on OCI’s unique abilities, so a new approach is warranted.

Numerous studies have explored machine learning techniques to extract cloud properties from satellite sensor data. These include both retrieval of a single cloud property~\cite{liu2022machine,wang2020machine,he2021dabnet}, and simultaneous retrievals of multiple cloud properties~\cite{wang2022retrieval,huang2022acm}. However, challenges persist in this area. To begin with, it's unclear how incorporating atmospheric domain knowledge, such as physical relationships between cloud properties, alongside advanced machine learning techniques, enhances retrieval accuracy. Furthermore, despite the deployment of numerous satellite sensors for similar cloud retrieval tasks (as discussed in Section~\ref{sec:definition}), the generalizability, especially from an Earth science perspective, of employing a unified machine learning architecture across different sensors remains uncertain. 

To address the above challenges we introduce MT-HCCAR, an end-to-end Multi-Task Learning (MTL) model for cloud masking, cloud phase prediction, and COT regression. Our contributions are fourfold. First, we incorporate hierarchical classification to capture the hierarchical relationship between cloud masking and cloud phase classification, enhancing prediction performance as demonstrated in the comparison experiment with baseline methods and ablation study. Second, we employ a cross-attention module to improve COT regression accuracy by leveraging similarities across the classification and regression networks. Third, we conduct quantitative experiments to demonstrate the advantages of our MT-HCCAR model over state-of-the-art baselines and its generalizability across three different satellite sensors. Finally, we perform quantitative and qualitative evaluation of performance variations across sensors from an Earth science perspective. Our implementation is publicly available at GitHub~\cite{Github-site}.

\section{Related Work}
\label{sec:relatedwork}

In recent years, various machine learning techniques, such as Random Forest (RF), Multi-Layer Perceptron (MLP), and Convolutional Neural Network (CNN), have been employed for retrieving different cloud properties. Among them, most of the work targets cloud detection~\cite{liu2022machine,wang2020machine}, cloud phase~\cite{guo2020cdnetv2}, and cloud thickness~\cite{shao2019cloud}. While these machine learning approaches for cloud property retrieval predominantly leverage spectral features, the main limitations lie in two aspects: 1) many studies do not consider background knowledge, such as the task order (e.g., cloud mask prediction, cloud phase prediction, and COT prediction), and 2) several studies conduct different tasks independently, lacking knowledge sharing between classification and regression tasks. These limitations underscore the need for more integrated and informed methodologies in cloud property retrieval studies.

Multi-task learning (MTL) is proving valuable in Earth science and remote sensing by jointly enhancing performance across diverse remote-sensing tasks through shared features. The studies have shown that classification and regression tasks can be implemented together to improve model performance~\cite{ilteralp2021deep,chen2022novel}. There are also studies conducting ML learning specifically for cloud property retrieval. Yang et al.~\cite{yang2022machine} developed an MLP-based method to retrieve cloud macrophysical parameters (cloud mask, cloud top temperature, and cloud top height) using Himawari-8 satellite data. Wang et al.~\cite{wang2022retrieval} proposed TIR-CNN based on the U-Net model to retrieve cloud properties including cloud mask, COT, effective particle radius (CER), and cloud top height (CTH) from thermal infrared radiometry. The architecture consists of encoding and decoding layers, convolutional blocks, batch normalization layers, and leaky Rectified Linear Units (ReLU). The results of applying the model to thermal infrared spectral data from MODIS are used to compare model performance for daytime and nighttime data. 

While these above studies offer valuable insights that are beneficial to our design, our work cannot be compared with most of them directly via experiments. This is because the OCI data we use does not have the same spectral bands, such as TIR bands, with those methods used. Consequently, we will not compare our model with architectures employing convolutional layers, such as CNN.

\begin{figure*}[btp]
\centering
\includegraphics[width=0.8\textwidth]{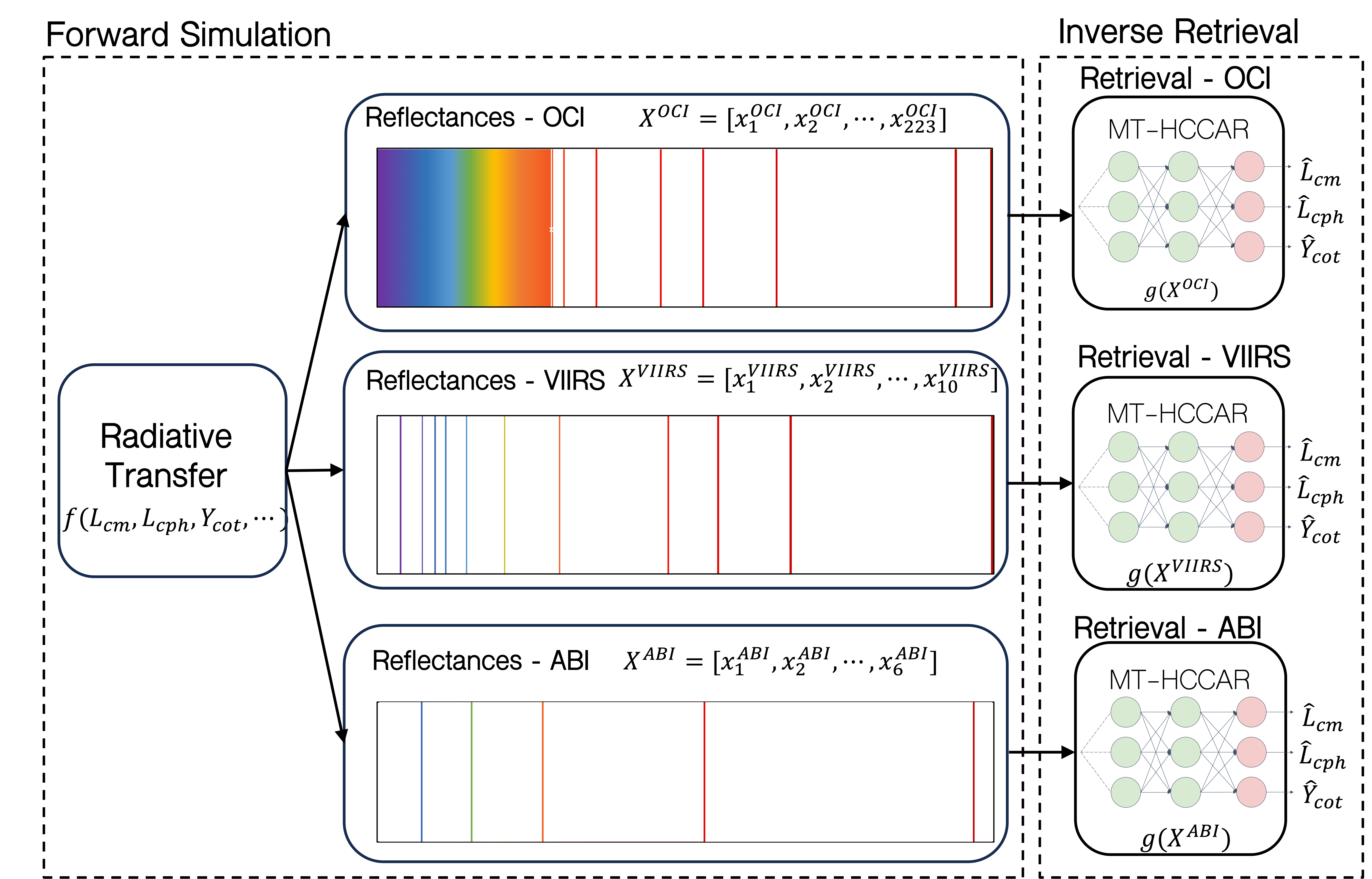}
\caption{Illustration of forward simulation and retrieval. $\mathbf{x_M}$: the spectral and physical features. $f(\cdot)$, $g(\cdot)$: the RT simulation algorithm and the retrieval algorithm. $\mathbf{L}_{cm}, \mathbf{L}_{cph}, \mathbf{Y}_{COT}$: the labels of cloud mask, cloud phase, and COT values, respectively.}
\label{fig:forwardbackward}
\end{figure*}

\section{Problem Statement and Data Simulation}
\label{sec:definition}

Cloud property determination requires two operations: simulation and retrieval. Simulation is a forward process mapping from the Earth's state (cloud and surface properties) to satellite direct observations (the reflected light signal the satellites see). Retrieval is an inverse process, determining the geophysical quantities from the satellite observations. Figure~\ref{fig:forwardbackward} illustrates these two processes with colors showing differences in the wavelength band of the three sensors.


\subsection{Radiative Transfer Simulation}
\label{sec:simulation}

The simulation process provides training data for our cloud property retrieval model. This study was conducted before the launch of PACE, and the spacecraft has not entered routine operations. Therefore, simulated datasets for OCI, VIIRS, and ABI were utilized to evaluate the sensors and architectures against a standardized source, enabling direct comparisons. This approach ensures consistency as the satellites are tested on identical simulated scenes with the same viewing geometries, which would not be feasible using real data.

Radiative transfer (RT) models describe the scattering and absorption processes that affect the propagation of light through Earth's atmosphere and surface. RT is the forward model to map from surface and atmospheric properties to spectral intensities seen by the space-borne sensors. We account for realistic variations in the properties of clouds (phase, microphysical, optical, and vertical structure), the atmospheric profile (aerosol, gas, Rayleigh scattering, temperature, and pressure), surface reflectance, and solar/observation geometry. Further details are as in Sayer et al.~\cite{sayer2023chroma}, with two modifications: 1) simulating cloud-free conditions as well as single-layer clouds, and 2) utilizing 20 different surface reflectance classes included in the libRadtran RT model~\cite{emde2016libradtran} (including various land surface spectra, water, and snow/sea ice). The simulation does not include multi-layer or mixed-phase cloud systems, aligning with most satellite retrieval processing algorithms. However, future work may extend the network to include such systems. We generate a dataset of 250,000 simulations for model training and evaluation. We simulate from the UV to SWIR and convolve these simulations with the solar spectrum from~\cite{coddington2021tsis} and sensor relative spectral response functions for OCI~\cite{OCI-site}, ABI~\cite{ABI-site}, and VIIRS~\cite{VIIRS-site} in order to generate the spectral top of atmosphere reflectance signal that the instruments would observe. This provides our simulated observations and reference truth (cloud classification and COT), along with band centers for each sensor. For \textbf{OCI}, there are 233 bands in total, including 226 hyperspectral bands ranging from 350 to 890 nm with 2.5 nm spacing, and seven discrete NIR/SWIR bands centered near 940, 1040, 1250, 1378, 1620, 2130, and 2260 nm. \textbf{VIIRS} has 10 bands centered near 412, 445, 488, 555, 672, 865, 1240, 1380, 1610, and 2250 nm. \textbf{ABI} has 6 bands centered near 471, 640, 860, 1370, 1600, and 2200 nm, respectively.

Each ABI band has a close analog in VIIRS, and similarly, each VIIRS band has a counterpart in OCI, thus forming a trio representing increasing complexity. Another commonly-used sensor, MODIS, has similar spectral coverage to VIIRS so we do not include it. Both ABI and VIIRS have TIR bands, but these are omitted from consideration as our primary focus is on developing a new methodology for OCI, which lacks TIR bands. Application to ABI and VIIRS demonstrates the broader applicability of our new architectures.

\subsection{Cloud Property Retrieval}
\label{sec:problem_definition}
Our primary objective is to accurately model the base-10 logarithm of COT (throughout the remaining text, COT means this logarithm unless explicitly described as the ``original'' COT) for pixels labeled as cloudy. We work in log space for COT as this has a more linear relationship with the brightness seen by satellite than the original COT. Simultaneously, the model should be able to accurately classify the cloud phase [cloudy, cloud-free, cloudy-liquid, cloudy-ice] for each pixel in the dataset to aid COT prediction. Therefore, our problem consists of two tasks: 1) a classification task to predict cloud mask and phase, and 2) a regression task to predict COT values. Assuming that the labels are $\bar{C} =$ `cloud-free', $C =$ `cloudy', $CL =$ `cloudy-liquid', $CI = $`cloudy-ice', and bold notations indicate arrays, the details of our study are outlined below. 

\textbf{Model input}. 1) \textbf{Input features} are represented as $\mathbf{X} = [\mathbf{x_1}, \mathbf{x_2}, \ldots, \mathbf{x_M}]$, where $M$ is the dimension of the available features. The exact input feature variables include: i) surface pressure in millibar (mbar), ii) total column water vapor content in millimeters (mm), iii) total column ozone content in Dobson units (DU), iv) types of Earth surface with 4 categories (land, snow, desert, and ocean/water), v) top of atmosphere reflectance at different wavelengths collected by each spaceborne sensor, vi) viewing zenith angle, solar zenith angle, and the relative azimuth angle, in degree. 2) \textbf{Cloud mask/phase labels} $\mathbf{l^{cls}}$ are used to train the classification task of the model. The set of possible label values is $l = \{C, \bar{C}, CL, CI\}$. There is a hierarchical relationship between the labels, as pixels covered by liquid cloud or ice cloud are both cloudy pixels. Also, there is no coexistence between liquid cloud and ice cloud in our data. Thus, $CL$ and $CI$ can be two subclasses of label $C =$`cloudy'. 3) \textbf{True COT value} $\mathbf{y^{cot}}$ are used to train the regression task of the model. COT values are not available for pixels with no cloud cover. That is, if ${l^{cls}_i=0}$, then $ y^{cot}_i = N/A$.

\textbf{Model output}. 1) Predicted cloud mask/phase class $\mathbf{\hat{l}^{cls}}$ with probabilities $\mathbf{u}$ of each pixel belonging to each of the four classes. 2) Predicted COT value $\mathbf{\hat{y}^{cot}}$ for each pixel. 3) Model architecture $\mathcal{M}$ with parameters $\mathbf{\beta}$. The predicted values are generated by the model, which is $[\mathbf{\hat{y}^{cot}}, \mathbf{\hat{l}}^{cls}] = \mathcal{M}(\mathbf{X|\beta})$.

\section{MT-HCCAR Model}
\label{sec:methods}


\begin{figure*}[btp]
\centering
\includegraphics[width=0.9\textwidth]{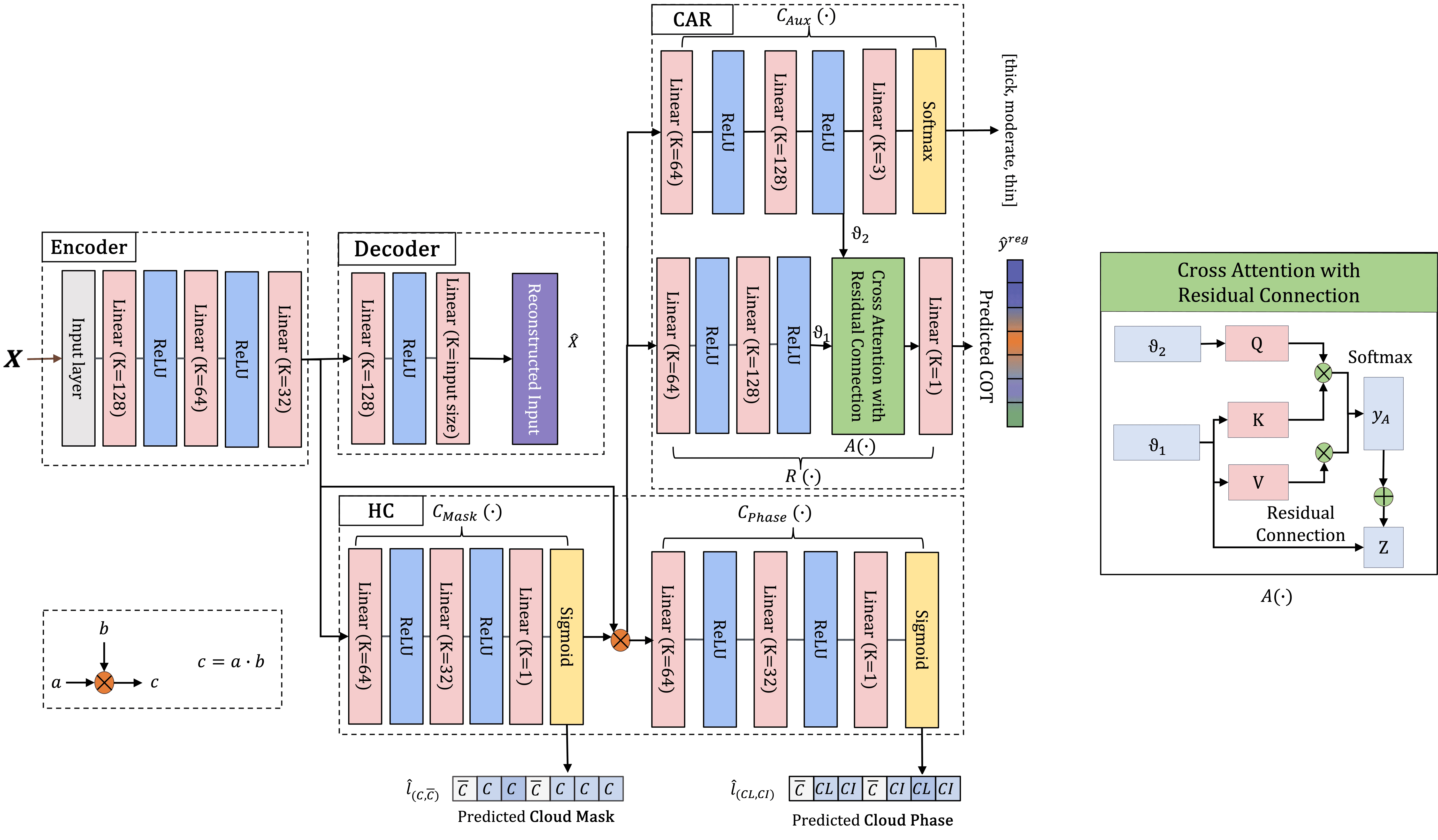}
\caption{MT-HCCAR: an end-to-end MTL model with hierarchical classification (HC) and cross attention assisted regression (CAR). The HC sub-network comprises the cloud masking module $C_{Mask}$ and the cloud phase classification module $C_{Phase}$. The CAR sub-network consists of the auxiliary coarse classification module $C_{Aux}$, the cross attention module $A$, and the regression module $R$. On the right is the structure of $A$.}
\label{fig:hccar}
\end{figure*}

Our objective is to train a deep learning model to accomplish two tasks: 1) the classification of cloud mask and phase for each pixel based on its reflectance values, and 2) the subsequent prediction of COT values for pixels classified as cloudy. Toward this objective, our proposed MT-HCCAR model is illustrated in Figure~\ref{fig:hccar} and the subsections below will each of the modules.

\subsection{Encoder-Decoder Sub-Network}
\label{sec:mtl}

An encoder-decoder sub-network, containing an encoder module and a decoder module, is employed in our model to learn latent feature parameters that can be shared for other learning tasks. To integrate the COT regression task and cloud mask/phase prediction task, we adopt the soft-parameter sharing approach of MTL where the shared parameters are derived by encoder-decoder. Previous studies~\cite{toshniwal2017multitask,kuga2017multi} have used similar encoder-decoder techniques to transform original features into more relevant features for better downstream prediction accuracy. 
 In our model, the decoder reconstructs the input feature $\mathbf{X}$ into $\mathbf{\hat{X}}$, and a loss is between $\mathbf{X}$ and $\mathbf{\hat{X}}$ is minimized during training to minimize the distortion of input feature throughout feature extraction layers.

The shared encoder is formed by the first three dense layers, each of which is followed by a ReLU activation function. This feature extractor is a wide-to-narrow structure where the number of kernels at each layer decreases gradually. 
The decoder and reconstruction branch is narrow-to-wide and further improves the performance of the encoder to learn features for different tasks.

\subsection{Hierarchical Classification (HC) Sub-Network}
\label{sec:mtl-hc}

Utilizing the shared parameters acquired by the encoder-decoder sub-network, MT-HCCAR incorporates a hierarchical classification sub-network. This sub-network comprises a cloud mask classification module and a cloud phase classification module, facilitating cloud mask and phase predictions. To enhance the physical interpretability of the model, the architecture is crafted to mirror human cognitive processes in understanding cloud labels. Fundamentally, to predict cloud phase and COT values from an Earth science perspective, the model must discern between cloudy and non-cloudy pixels before further classifying liquid-phase and ice-phase pixels and predicting COT values for cloudy pixels.

Hierarchical classification has the capability to categorize instances according to label levels, forming a tree-like structure where each label functions as a node on the tree~\cite{miranda2023hiclass}. The HC network used in MT-HCCAR consists of two classifiers $C_{Mask}$ and ${C_{Phase}}$, the first of which distinguishes cloudy pixels from the cloud-free pixels, and the output is uncertainties for the two labels $[\mathbf{\hat{u}^{C}}, \mathbf{\hat{u}^{\bar{C}}}]$. The output uncertainties are binarized to $[\mathbf{\hat{l}^{C}}, \mathbf{\hat{l}^{\bar{C}}}]$ by a threshold of 0.5. Before the beginning of the second classifier ${C_{Phase}}$ input feature $\mathbf{X}$ is multiplied by the predicted cloud label, $\mathbf{\hat{y}^{C}}$, to filter out cloud-free pixels. Then ${C_{Phase}}$ classifies liquid-cloud and ice-cloud and the sigmoid uncertainties $[\mathbf{\hat{u}^{CL}}, \mathbf{\hat{u}^{CI}}]$ are transformed in the same way to binary cloud-phase labels $[\mathbf{\hat{l}^{CL}}, \mathbf{\hat{l}^{CI}}]$. 

\subsection{Classification Assisted Regression Sub-Network based on Cross Attention Mechanism (CAR)}
\label{sec:car}

Taking inputs from the encoded feature set and cloud mask classification results as input, our regression sub-network CAR is used for further downstream regression prediction of a cloudy pixel's COT value. 
Two novel efforts were taken to improve COT prediction accuracy. First, instead of a direct regression module, an auxiliary coarse classification module is added to predict which sub-range of COT each pixel falls into. Second, a residual-based cross-attention mechanism inspired by~\cite{vaswani2017attention} is developed to enable the regression module and the auxiliary classification module to share relevant correlations and insights.

As illustrated in Figure~\ref{fig:hccar}, there is a connection between internal features from the auxiliary classifier and those from the regression network, facilitated by a residual-based cross-attention module. The close alignment in tasks between the auxiliary classifier and the regression network enables the cross-attention mechanism to selectively attend to features relevant to one task while simultaneously executing the other. This contrasts with relationships involving the regression task and other tasks, such as cloud phase classification, where associations are forged through joint optimization of their respective losses during training iterations, rather than through the shared utilization of internal features.

Our auxiliary coarse classification involves discretizing continuous values into coarse groups, serving as a preprocessing step before regression to align pixels with similar characteristics in the feature space. Moreover, the auxiliary classifier assigns greater importance to the regression task during model training to get accurate COT predictions. In MT-HCCAR, an auxiliary classifier is employed to categorize continuous COT values into three distinct levels: thin cloud, moderate cloud, and thick cloud. Specifically, we define thin cloud pixels as those with logarithmic COT values within the range of [-1.5, 0], moderate cloud in [0, 1], and thick cloud in [1.0, 2.5]. Given the analogous nature of the tasks in COT coarse classification and COT regression, the features extracted by the auxiliary classifier are integrated with those obtained from the regression network.

To learn the joint features more effectively, we utilize cross-attention layers to enhance the integration of features from both regression layers and coarse classification layers, facilitating deeper feature learning and more cohesive feature amalgamation. Based on the cross-attention mechanism, adding a residual connection guarantees that the module has a stable output and related research such as Wang et al.~\cite{wang2018non} propose a non-local operation network using a residual connection to insert blocks to the network. By using the residual connection in our module $A$, the output is $A(\vartheta_1, \vartheta_2)= W_z y_{A}(\vartheta_1, \vartheta_2) + \vartheta_1$,  where as shown in Figure~\ref{fig:hccar} $\vartheta_1$ is the feature from the regression network, $\vartheta_2$ is the feature from the auxiliary classification module, $W_z$ is the weight matrix to map $y_{A}$ to $\vartheta_2$, and $y_{A}$ denotes the computation within cross-attention mechanism. The calculation of $y_{A}$ involves three variables: a query ($Q$), a key ($K$) and a value ($V$)~\cite{vaswani2017attention}. They are calculated based on corresponding weights matrices $W_Q$, $W_K$, $W_V$ learned through model training. $Q = W_Q \vartheta_2$, $K = W_K \vartheta_1$, and $V = W_V \vartheta_1$. Then the $y_{A}$ is $y_{A} = Softmax(QK^T)V$.

\subsection {Model Training of MT-HCCAR}
\label{sec:training}

The loss function $\mathcal{L}_{MT-HCCAR}$ for training MT-HCCAR model is formulated as the weighted sum of four components: a hierarchical classification loss $\mathcal{L}_{HC}$, a regression loss $\mathcal{L}_{CAR}$, a reconstruction loss $\mathcal{L}_{Rec}$, and a Lasso regularization loss $\mathcal{L}_{Lasso}$. That is: $\mathcal{L} = \mathcal{L}_{HC}+\mathcal{L}_{CAR}+\mathcal{L}_{Rec}+\mathcal{L}_{Lasso}$. The four loss components in the loss function are calculated using different rules.

Among the four components, The reconstruction loss $\mathcal{L}_{Rec}$ and the Lasso regularization loss component $\cal{L}_{Lasso}$ can be directly calculated.
The reconstruction loss $\mathcal{L}_{Rec}$ describes the difference between the input features $\mathbf{X}$ and the reconstruction of the input generated through the encoder and the decoder. The loss function is a mean square error (MSE) between $\mathbf{D}(\mathbf{X}) = \mathbf{\hat{X}}$ and $\mathbf{X}$, which means $\mathcal{L}_{Rec} = \sum_{i=1}^{n}(\mathbf{X_i} - \mathbf{\hat{X_i}})^2$. The Lasso regularization loss $\mathcal{L}_{Lasso}$ is an additional penalty to regularize the training process by increasing the sparsity of the model with the Lasso regularization loss. $ \mathcal{L}_{Lasso} = \lambda\sum_{p=1}^{P}|\beta_p|$.

The other two components, HC loss and CAR loss, are both weighted sums of losses from the supplement modules. The HC loss is the summation of Binary Cross Entropy loss from cloud masking module $C_{Mask}$ and cloud phase classification module $C_{Phase}$. The HC loss function is $\mathcal{L}_{HC} = \mathcal{L}_{C_{Mask}} + \mathcal{L}_{C_{Phase}}$. We have a sigmoid output of cloudy and non-cloud uncertainties $[\hat{u}^{C}, \hat{u}^{\bar{C}}]$ from cloud mask classifier $C_{Mask}$ and sigmoid output of liquid cloud and ice cloud uncertainties $[\hat{u}^{CL}, \hat{u}^{CI}]$ from cloud phase classifier $C_{Phase}$, then for the HC loss, $\mathcal{L}_{C_{Mask}}$ and $\mathcal{L}_{C_{Phase}}$ are calculated as $\mathcal{L}_{C_{Mask}} = -\frac{1}{N}\sum_{i=1}^{N} l^{C}_i\cdot \log(\hat{u}^C_i) + l^{\bar{C}}_i\cdot \log(\hat{u}^{\bar{C}}_i)$ and $\mathcal{L}_{C_{Phase}} = -\frac{1}{N}\sum_{i=1}^{N} \hat{u}^C_i l^{CL}_i\cdot \log(\hat{u}^C_i \hat{u}^{CL}_i) + \hat{u}^C_i l^{CI}_i\cdot \log( \hat{u}^C_i \hat{u}^{CI}_i)$.

\noindent The CAR loss $\mathcal{L}_{CAR}$ is the summation of Cross Entropy loss from the auxiliary classifier $C_{Aux}$ and $l1$ loss from regression. That is, $\mathcal{L}_{CAR} = \mathcal{L}_{R}+ \mathcal{L}_{C_{Aux}}$. We have predicted COT values and $\hat{y}^{cot}$ and sigmoid output of thickness group uncertainties $[\hat{u}^{thin}, \hat{u}^{mod}, \hat{u}^{thick}]$ from the auxiliary classifier, then the two components are $\mathcal{L}_{R} = \sum_{i=1}^{N}|{y}^{cot}_i- \hat{y}^{cot}_i |$ and $\mathcal{L}_{C_{Aux}} = -\sum_{i=1}^{N}\sum_{c\in[thin, mod, thick]} l^{c}_i \ln{\hat{u}^{c}_i}$.


\section{Experiments}
\label{sec:experiments}

 To evaluate our model's performance, we compare MT-HCCAR with baseline methods and conduct an ablation study across OCI, VIIRS, and ABI datasets. 
 
 Given that our dataset comprises independent pixels, we choose two baseline methods from prior research with comparable task objectives and data formats for our experiments: 1) \textbf{Chen et al. (2018)}~\cite{Chen2018}: An MLP based method for cloud property retrieval. We use 1 hidden layer with 10 nodes, which are the same as the authors used. 2) \textbf{Liu et al. (2022)}~\cite{liu2022machine}: An RF based method for cloud detection. We apply one RF to cloud masking and the same RF to cloud phase classification, respectively. The parameters involved in the method are $ntrees=100$ and $mdepth=10$.

 Besides the above two baseline methods, four ablation study models are also used to evaluate the effectiveness of different modules in our model. The comparison of the four models and our MT-HCCAR is shown in Table~\ref{tab:com}. The detailed differences can be found in the supplementary material.

\begin{table*}[ht]
    \centering
    \caption{Comparison of models in the ablation study. STL: single-task learning.}
    \begin{tabular}{llll}
    \toprule
         \textsc{Model} & \textsc{Structure}  & \textsc{Classification} & \textsc{Regression}  \\
    \midrule
         SEQ & STL & Regular & Regular \\
         MT-CR & MTL & Regular & Regular \\
         MT-HCR & MTL & HC & Regular \\
         MT-HCCR & MTL & HC & CAR without Attn. \\
         \textbf{MT-HCCAR}& \textbf{MTL} & \textbf{HC} & \textbf{CAR} \\
    \bottomrule
    \end{tabular}
    \label{tab:com}
\end{table*}


\subsection{Experiment Setup}
\label{sec:expset}

Our model was implemented using the Python deep-learning library PyTorch. All baseline models and proposed models were tested on a single GPU on Kaggle. The dataset comprises satellite data simulation of instruments OCI, VIIRS, and ABI. The three datasets encompassing $N = 250,000$ instances. 

We split data into 62.5\%, 22.5\%, and 10.0\% for training, validation, and test sets, respectively. All models are trained with the same hyperparameters including learning rate = $1e^{-5}$, batch size = $64$, and training epochs = $500$. All experiment results are mean values of metrics from a 10-fold cross-validation. 

 Simulations of OCI, VIIRS, and ABI are utilized instead of observations as our labeled training dataset principally because there is no suitable comprehensive reference truth dataset. While the standard retrieval products from these sensors could be used to train networks, for example, these products have known limitations so there would be a risk of the model training to the artifacts in these products. In a similar vein, comparing the results of applying the trained NN to real satellite observations of these standard products will be instructive to get a general sense of reasonableness but not to gauge their absolute performance. It is expected that, by using a more realistic training set of simulations than was used to develop the physically-based retrievals (which include many simplifying assumptions out of the computational necessities from decades ago when these approaches were developed), the NN model should outperform them.

\subsection{Evaluation Metrics}
\label{sec:metrics}
We use three types of metrics to evaluate our work, including metrics for the classification task, metrics for the regression task, and metrics from an Earth science perspective.

\textbf{Metrics for classification performance}. We use Accuracy and Average precision to evaluate classification performance. Cloud masking accuracy (ACC$_{bi}$) is the fraction of correct predictions of two big categories [cloudy, cloud-free] by our model. The area under the precision-recall curve ($AU(\overline{PRC})_c$) is calculated for each label. The weighted area under the precision-recall curve ({$AU(\overline{PRC})_w$}) is the weighted mean of precisions across all labels at each threshold $h$ for class $i$, with the weight as subtracted between recall at threshold $h$ and the recall at threshold $h-1$, where the number of thresholds is close to infinity. The weighted precision and recall are: $R^h_w = \frac{\sum_{l}TP^h_c}{\sum_{l}TP^h_c+\sum_{l}FN^h_c}$ and $P^h_w = \frac{\sum_{l}TP^h_c}{\sum_{l}TP^h_c+\sum_{l}FP^h_c}$, then $AU(\overline{PRC})_w$ is: $AU(\overline{PRC})_w = \sum_{h}(R^h_w  - R ^{h-1}_w)P^h_w$.



\textbf{Metrics for regression performance}. We use MSE and $R^2$ to evaluate regression performance. Mean squared error (MSE) is the average squared difference between the true and predicted COT values. $MSE = \sum_{i=0}^{n}(y_i^{cot} - \hat{y}_i^{cot})$. 
The coefficient of determination \textbf{($R^2$)} measures how close the predicted value is to the true value. $R^2 = 1- \frac{\sum_{i=0}^{n}(y_i^{cot} - \hat{y}_i^{cot})^2}{\sum_{i=0}^{n}(y_i^{cot} - \overline{y}^{cot})^2}$, where $\overline{y}^{cot}$ is the mean value of all true COT values $[y_1^{reg}, y_2^{reg}, \ldots, y_n^{reg}]$.

\textbf{Earth science metrics}. Besides the above metrics, we also evaluate how a machine learning model could be used for actual satellite missions. The fraction of pixels meeting PACE goals (FMG) is an evaluation metric defined by the PACE Validation Plan~\cite{PACE-VP-site} based on scientific requirements and expectations. For each pixel, the  relative error $Error_i = |\frac{y^{(cot)_i} - \hat{y}^{(cot)_i} }{y^{(cot)_i}}|$. Then FMG represents the percentage of pixels whose relative error is less than 0.25 (for liquid clouds) or 0.35 (for ice clouds). The Validation Plan defines a predictive model with satisfactory performance as one where this goal is met for 65\% of cloudy pixels. Note that this mission goal only applies to cases where the true original COT value is 5 or more (log10 COT \textgreater{} 0.7). 

\subsection{Comparison with Baseline Models}
\label{sec:baseline}

Tables \ref{tab:cls} and \ref{tab:reg} present a comprehensive evaluation of our proposed model MT-HCCAR against two baseline methods, demonstrating superior performance across both classification and regression metrics. Specifically, in the assessment of classification tasks, ACC$_{bi}$ pertains to cloud masking performance, while $AU(\overline{PRC})_w$ and $AU(\overline{PRC})_c$ quantify the performance of both cloud masking and cloud phase classification. 

Upon close examination of Table \ref{tab:cls}, the RF classifier by \cite{liu2022machine} attains superior ACC$_{bi}$, $AU(\overline{PRC})_w$, and $AU(\overline{PRC})_c$ in comparison to the SEQ model and the simplest MTL based model, MT-CR, within the ablation study. However, MTL-based models featuring the HC module including MT-HCR, MT-HCCR, and MT-HCCAR, surpass the performance of the RF classifier in these metrics. Shifting the focus to the results of the COT regression task in Table \ref{tab:reg}, the MLP-based baseline method introduced by \cite{Chen2018} produces results on par with the SEQ model but significantly inferior to MTL-based models. This finding emphasizes the effectiveness of MTL-based models for COT regression.

\begin{table*}[ht]
    \centering
    \caption{Model performance comparison with baseline models and ablation study - classification results. Models are trained on OCI.}
    \begin{tabular}{p{2.0cm}p{1.4cm}lp{1.5cm}llp{1.4cm}}
    \toprule
         \textsc{Model} & \textsc{ACC$_{bi}$}  & \textsc{$AU(\overline{PRC})_w$} & \multicolumn{4}{c}{$AU(\overline{PRC})_c$} \\
         	&		&		&	Cloudy	&	Cloud-free	&	Liquid cloud	&	Ice cloud \\
        \midrule
            ~\cite{liu2022machine} & 0.968 & 0.985 &   0.955 & 0.949 & 0.951 & 0.965\\
            ~\cite{Chen2018} & 0.942 & 0.638 & 0.974 & 0.882 & 0.385 & 0.348 \\
            \midrule
            \midrule
            SEQ	&	0.963	&	0.750	&	0.997	&	0.978	&	0.618	&	0.658	\\
            MT-CR	&	0.965	&	0.955	&	0.925	&	0.925	&	0.954	&	0.967	\\
            MT-HCR	&	0.975	&	0.993	&	0.999	&	0.989	&	0.982	&	0.992	\\
            MT-HCCR	&	0.982	&	0.995	&	0.999	&	0.992	&	0.985	&	0.993	\\
            MT-HCCAR	&	\textbf{0.984}	&	\textbf{0.996}	&	\textbf{0.999}	&	\textbf{0.995}	&	\textbf{0.990}	&	\textbf{0.996}	\\
    \bottomrule
    \end{tabular}
    \label{tab:cls}
\end{table*}

\begin{table*}[h]
    \centering
    \caption{Model performance comparison with baseline models and ablation study - regression results. Models are trained on OCI. The model~\cite{liu2022machine} is not applicable for the regression task.}
    \begin{tabular}{lp{1.1cm}p{1.3cm}p{1.3cm}p{1.1cm}p{1.1cm}p{1.1cm}p{1.3cm}p{1.3cm}}
         \toprule
\textsc{Model}& \textsc{MSE} & \multicolumn{2}{c}{\textsc{MSE each label}} & \textsc{$R^2$} & \multicolumn{2}{c}{\textsc{$R^2$ each label}} & \multicolumn{2}{c}{\textsc{FMG}} \\
	&		&	Liquid	&	Ice	&		&	Liquid	&	Ice	&	Liquid	&	Ice	\\
\midrule
~\cite{Chen2018} &	0.185	&	0.193	&	0.178	&	0.235	&	0.247	&	0.212	&	43.92\%	&	60.64\%	\\
\midrule
\midrule
SEQ	&	0.126	&	0.133	&	0.118	&	0.440	&	0.274	&	0.548	&	4.00\%	&	41.11\%	\\
MT-CR	&	0.062	&	0.075	&	0.049	&	0.728	&	0.661	&	0.788	&	44.46\%	&	68.21\%	\\
MT-HCR	&	0.038	&	0.045	&	0.030	&	0.843	&	0.810	&	0.874	&	65.93\%	&	82.28\%	\\
MT-HCCR	&	0.031	&	0.035	&	0.028	&	0.869	&	0.854	&	0.884	&	66.38\%	&	82.43\%	\\
MT-HCCAR	&	\textbf{0.026}	&	\textbf{0.032}	&	\textbf{0.021}	&	\textbf{0.891}	&	\textbf{0.869}	&	\textbf{0.914}	&	\textbf{67.96\%}	&	\textbf{82.70\%}	\\
\bottomrule
    \end{tabular}
    \label{tab:reg}
\end{table*}

\begin{table}[ht]
    \centering
     \caption{Model performance for VIIRS and ABI datasets.}
     \label{tab:results}
     \begin{tabular}{p{1.5cm}p{2.1cm}p{1.2cm}p{1.6cm}p{1.2cm}p{1.2cm}p{1.3cm}p{1.3cm}}
     \toprule
         \textsc{Dataset}	&	\textsc{Model}	&	ACC$_{bi}$	&	$AU(PRC)_w$	& MSE  & $R^2$ & \multicolumn{2}{l}{FMG ($CL$, $CI$)}  \\
         \midrule
            \multirow{3}{*}{\textsc{VIIRS}}	
&	MT-CR	&	0.976	&	0.939	&	0.043	&	0.810 	&	48.52\% & 72.73\%	\\										
&	MT-HCR	&	0.989	&	0.996	&	0.030	&	0.874	&	66.23\% & 82.62\%	\\	
&	MT-HCCAR	&	\textbf{0.989}	&	\textbf{0.997}	&	\textbf{0.025 }	&		\textbf{0.897}	&	\textbf{68.20\%} & \textbf{82.69\%}	\\		\hline	
            \multirow{3}{*}{\textsc{ABI}}	
&	MT-CR	&	0.973	&	0.969	&	0.057	&		0.751	&	47.87\% & 71.81\%	\\		
&	MT-HCR	&	0.986	&	\textbf{0.996}	&	0.038	&	0.840	&	66.58\% & 82.55\%	\\
&	MT-HCCAR	&	\textbf{0.987}	&	\textbf{0.996}	&	\textbf{0.032}	&	\textbf{0.866}	&	\textbf{68.54\%} & \textbf{83.34\%}	\\			
     \bottomrule
     \end{tabular}
 \end{table}

\subsection{Ablation Study}
\label{sec:ablation}

The ablation study with five models from SEQ to MT-HCCAR for OCI, as depicted in the lower part of Tables~\ref{tab:cls} and~\ref{tab:reg}, highlights the effectiveness of our model in improving both classification and regression performance. The comparative analysis demonstrates significant improvements across all metrics from SEQ to MT-CR, which shows the usefulness of the MTL structure. For MTL methods, MT-HCR outperforms MT-CR in binary classification (cloudy, cloud-free) and refines liquid and ice cloud phase classification, leading to enhancements in regression metrics. This proves the effectiveness of the HC module. Moreover, despite focusing on regression, the introduction of CAR in MT-HCCAR maintains the performance of classification tasks. The efficacy of the CAR module is further validated by comparing MT-HCCAR to MT-HCR, indicating improvements in all metrics. Additionally, the adoption of the cross-attention module is substantiated by superior performance in MT-HCCR compared to MT-HCCAR, emphasizing its role in facilitating information exchange between hierarchical classification and regression. 

To confirm the generalizability of our proposed model, we did a further ablation study with VIIRS and ABI. As shown in Table~\ref{tab:results}, the application of MTL-based models reveals a similar trend of performance enhancement when integrating HC and CAR modules. Notably, MT-HCCAR achieves the highest performance across both classification and regression tasks among the MTL models across OCI, VIIRS, and ABI datasets, underscoring the generalization capabilities of the introduced model MT-HCCAR. The detailed results of cross-validation and model selection are in the supplementary material.

\begin{figure}[h]
\centering
\includegraphics[width=0.9\textwidth]{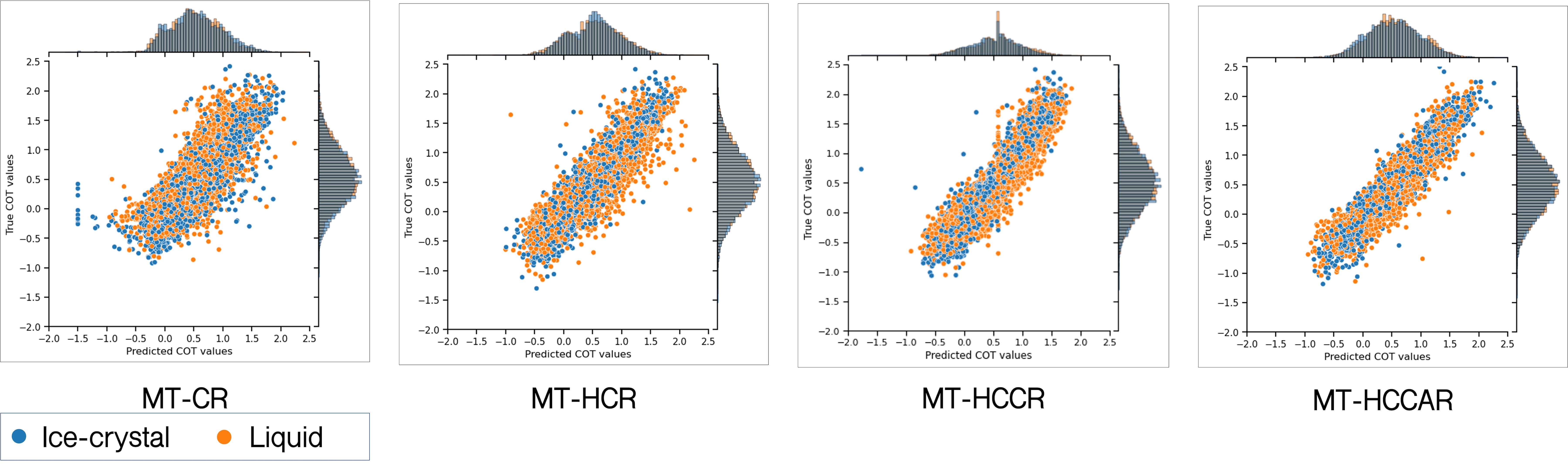}
\caption{Scatter plot with PDF distribution of true COT values (y-axis) and predicted COT values (x-axis) for the OCI dataset. }
\label{fig:scatter}
\end{figure}

To visually illustrate the improvements facilitated by the HC and CAR modules, Figure \ref{fig:scatter} presents a scatter plot showing the distribution of true COT values against predicted COT values for all instances in the test set. Integration of HC and CAR modules results in more accurate predictions compared to MT-CR and MT-HCR. Notably, the predicted Probability Density Function (PDF) aligns closely with the true distribution, indicating improved fidelity to actual COT values. Additionally, scatter points cluster more tightly along the diagonal line, confirming the model's enhanced precision when HC and CAR modules are incorporated into the model.

\subsection{Earth Science Evaluation}
\label{sec:results1}

\textbf{Quality of performance.} Satellite cloud masks are generally evaluated either through comparison against ground-based, airborne, or spaceborne observations from radar/lidar sensors that have much greater sensitivity for cloud detection than imagers in question. Accuracies tend to depend on the surface type and illumination conditions; for daytime scenes such as those simulated here, reported accuracies for NASA’s widely-used MODIS cloud mask are 0.850, 0.778, and 0.894. Chen et al.~\cite{Chen2018} developed an NN cloud mask for MODIS based on RT simulations and found an accuracy of 0.985 on training data but lower accuracies in the range of 0.739-0.885 on real data depending on region and time of year. Although real-world conditions are more complex than a simulation, the results suggest our models are competitive with current approaches. Note, that the cloud phase is less commonly validated in this way.

Obtaining a validation-grade measurement of COT presents significant challenges, with reference datasets lacking comprehensive coverage~\cite{platnick2020nasa}. Consequently, existing comparisons primarily focus on evaluating agreement among different remotely sensed datasets. Our network's performance suggests its potential to fulfill the objectives outlined in the Validation Plan~\cite{PACE-VP-site}. Notably, PACE's allowance for larger uncertainty in ice cloud measurements stems from the anticipated difficulties arising from variations in ice crystal properties in real-world scenarios. However, our results show that the uncertainty in ice COT can be smaller than that for liquid COT. Traditional methods often assume specific ice crystal properties, resulting in increased uncertainty in retrieved ice COT. Our findings indicate that satellite measurements inherently contain information about these properties, which neural networks are proficient at learning.

\textbf{Performance of different satellite sensors.} We also compared how the models perform for different satellite sensors based on the results in Tables~\ref{tab:cls},~\ref{tab:reg} and~\ref{tab:results}. The classification tasks exhibit high accuracy across all sensors, with virtually indistinguishable performance from a scientific perspective. The fundamental nature of cloud masking in Earth science prompts many sensors to incorporate a common set of bands proven to be effective for this purpose, with additional bands often designed for diverse applications. For instance, OCI's hyperspectral bands support the measurement of different ocean plankton species~\cite{Werdell2019}, revealing subtle spectral differences not readily discernible in multispectral data. In the regression task, OCI and VIIRS outperform ABI, indicating the utility of additional bands in predicting COT. The unexpectedly slightly superior performance of VIIRS over OCI may suggest that OCI's substantially larger feature space makes finding an optimal solution during training more challenging or could be attributed to stochastic variation. Adding training epochs for OCI may lead to better results, considering the number of bands.

\section{Conclusions}
\label{sec:conclusions}

In this study, we present MT-HCCAR, an end-to-end multi-task learning model tailored for cloud property retrieval on a simulated OCI satellite dataset in the PACE project, tackling tasks including cloud masking, cloud phase classification, and COT prediction. The model is implemented on three sensors' simulated datasets (OCI, VIIRS, and ABI), respectively, to examine its generalization. Comparative analyses against two baseline methods and ablation studies underscore the effectiveness of the HC module and the CAR module, enhancing performance in both classification and regression tasks. The ablation study establishes MT-HCCAR's superior performance across different datasets and multiple evaluation metrics. The positive results affirm our model's capability to address real-world challenges in cloud property retrieval and other multi-task applications. Future research endeavors will involve applying the model to spatial or temporal OCI images, co-located in space and time with VIIRS and ABI post-PACE launch, to assess its performance, and consistency, and enable detailed comparisons with deep learning models and non-machine-learning approaches.

\begin{credits}
\subsubsection{\ackname} This work is supported by a student fellowship from Goddard Earth Sciences Technology and Research (GESTAR) II, UMBC, grant OAC--1942714 from the National Science Foundation (NSF) and grant 80NSSC21M0027 from the National Aeronautics and Space Administration (NASA).

\end{credits}
%
%
%
\bibliographystyle{splncs04}
\bibliography{paper}

\appendix
\begin{center}
\textbf{\large Supplementary Materials - MT-HCCAR: Multi-Task Deep Learning with Hierarchical Classification and Attention-based Regression for Cloud Property Retrieval}
\end{center}


\author{Xingyan Li\inst{1}\orcidID{0009-0001-2598-2296} \and
Andrew M. Sayer\inst{2,3}\orcidID{0000-0001-9149-1789} \and
Ian T. Carroll\inst{2,3} \and 
Xin Huang\inst{4} \and
Jianwu Wang\inst{1}\orcidID{0000-0002-9933-1170}\corr}


\institute{Department of Information Systems, University of Maryland, Baltimore County, Baltimore, MD, USA
\email{\{xingyanli,jianwuw\}@umbc.edu}
\and
NASA Goddard Space Flight Center, Greenbelt, MD, USA 
\and
Goddard Earth Sciences Technology and Research (GESTAR) II, UMBC, Baltimore, MD, USA
\and
Dept of Computer \& Information Sciences, Towson University, Towson, MD, USA
}

\begin{abstract}
This additional material offers supplementary details of experiments discussed in the main paper published in ECML PKDD 2024.
\begin{itemize}
    \item Section A: Experiment Settings of Ablation Study (additional discussion to support the beginning of Section 5 and Table 1 in the main paper).
    \item Section B: Cross-Validation and Model Selection (additional discussion supporting Section 5.4 in the main paper).
\end{itemize}
\end{abstract}

\section{Experiment Settings of Ablation Study}

Four ablation study models are also used to evaluate the effectiveness of different modules in our model. The four models are 1) \textbf{SEQ}: A sequential structure model consists of three separate networks. The first one classifies [cloudy, cloud-free], the second one classifies [liquid-cloud, ice-cloud], and the third network makes regressive predictions on COT for cloudy pixels, respectively. 2) \textbf{MT-CR}: The MTL method combines the classification network and regression network in an end-to-end model. 3) \textbf{MT-HCR}: MTL model with the HC network classification task and a regression network in the MTL approach. 4) \textbf{MT-HCCR}: MTL model with hierarchical classification and classification-assisted regression, where only the cross-attention module is omitted from MT-HCCAR.

The purpose of comparing the four models with MT-HCCAR is to prove the effectiveness of multi-task learning (MTL), the effectiveness of the hierarchical classification sub-network (HC), the effectiveness of the auxiliary classification module in the CAR sub-network, and the effectiveness of the cross-attention module in the CAR sub-network, respectively. As described in the main manuscript, the objectives of this work include 1) classification of cloud masks (cloud existence), 2) classification of ice-cloud and liquid-cloud for cloudy pixels, and 3) regression of COT for cloudy pixels. Therefore, both classification networks and regression networks are necessary for whatever model we use to complete the three tasks. The difference lies in how we combine the networks, and what exact method we use to improve the performance of each task.

\begin{figure}[h]
\centering
\includegraphics[width=01.0\textwidth]{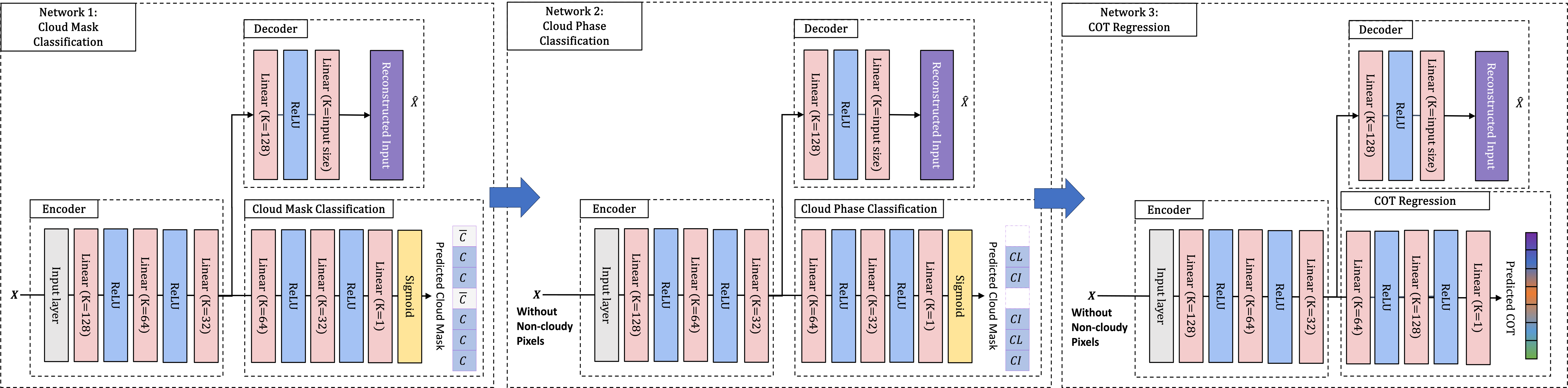}
\caption{The model SEQ, where SEQ is the abbreviation of Sequential Model. MTL method is not used for SEQ.}
\label{fig:seq}
\end{figure}

\begin{figure}[h]
\centering
\includegraphics[width=0.8\textwidth]{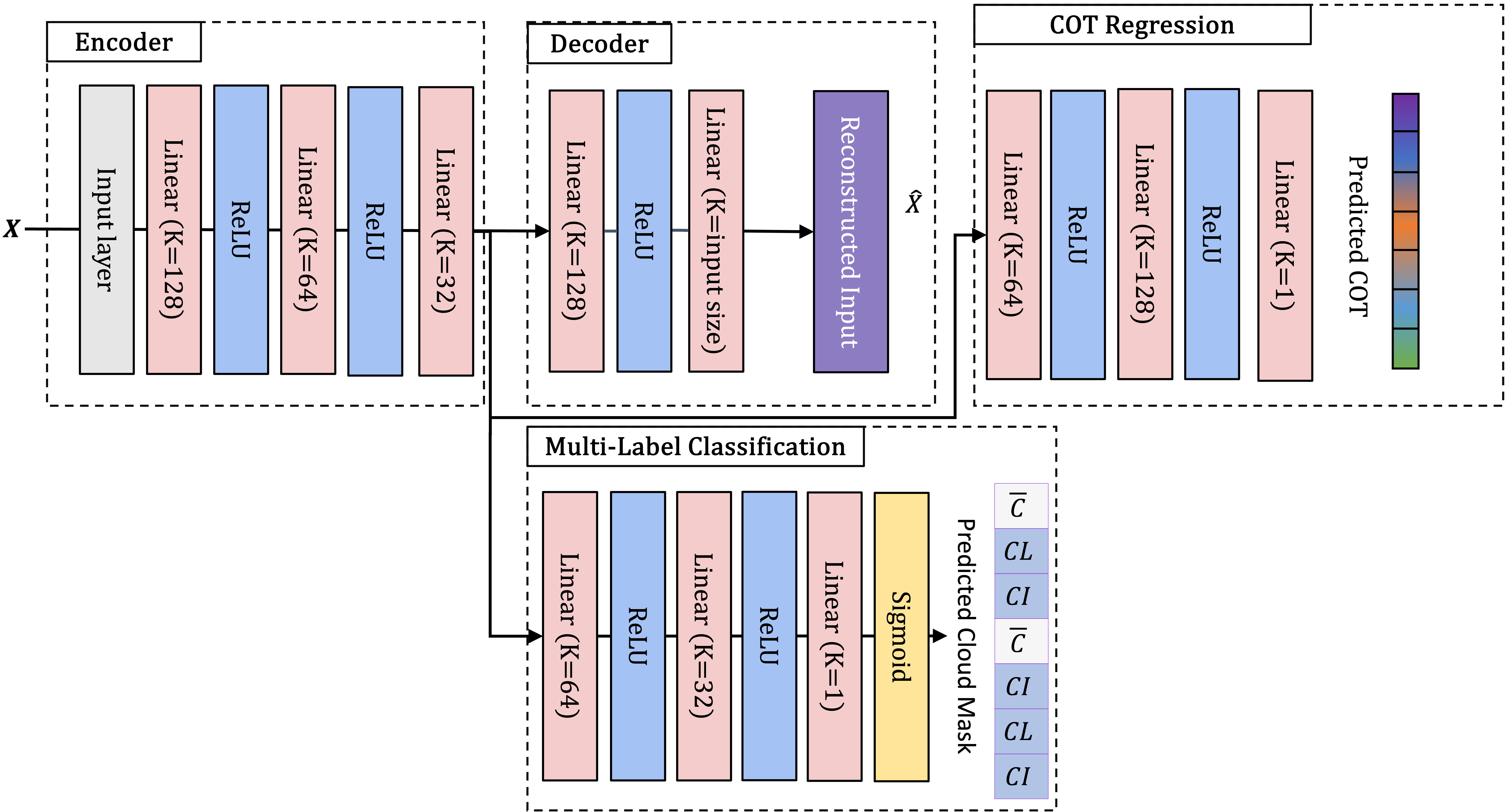}
\caption{The model MT-CR, where MT is the abbreviation of MTL, C represents Classification and R represents Regression. MTL method is used, but neither HC nor CAR sub-networks are used.}
\label{fig:mtcr}
\end{figure}

\begin{figure}[h]
\centering
\includegraphics[width=0.8\textwidth]{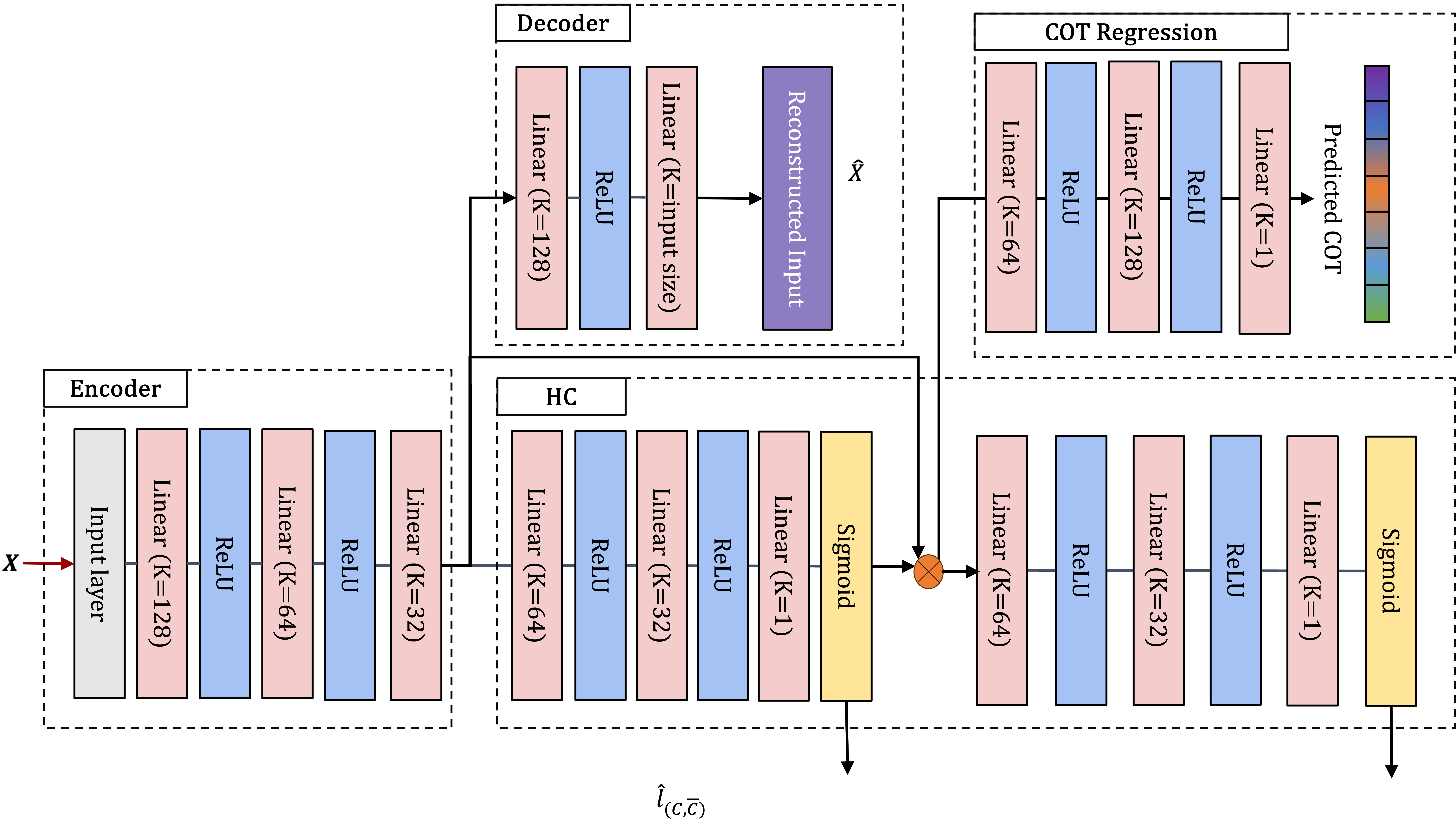}
\caption{The model MT-HCR, where HC represents the proposed HC sub-network and R represents Regression. MTL method and HC are used, while the CAR sub-network is not.}
\label{fig:mthcr}
\end{figure}

\begin{figure}[!h]
\centering
\includegraphics[width=0.8\textwidth]{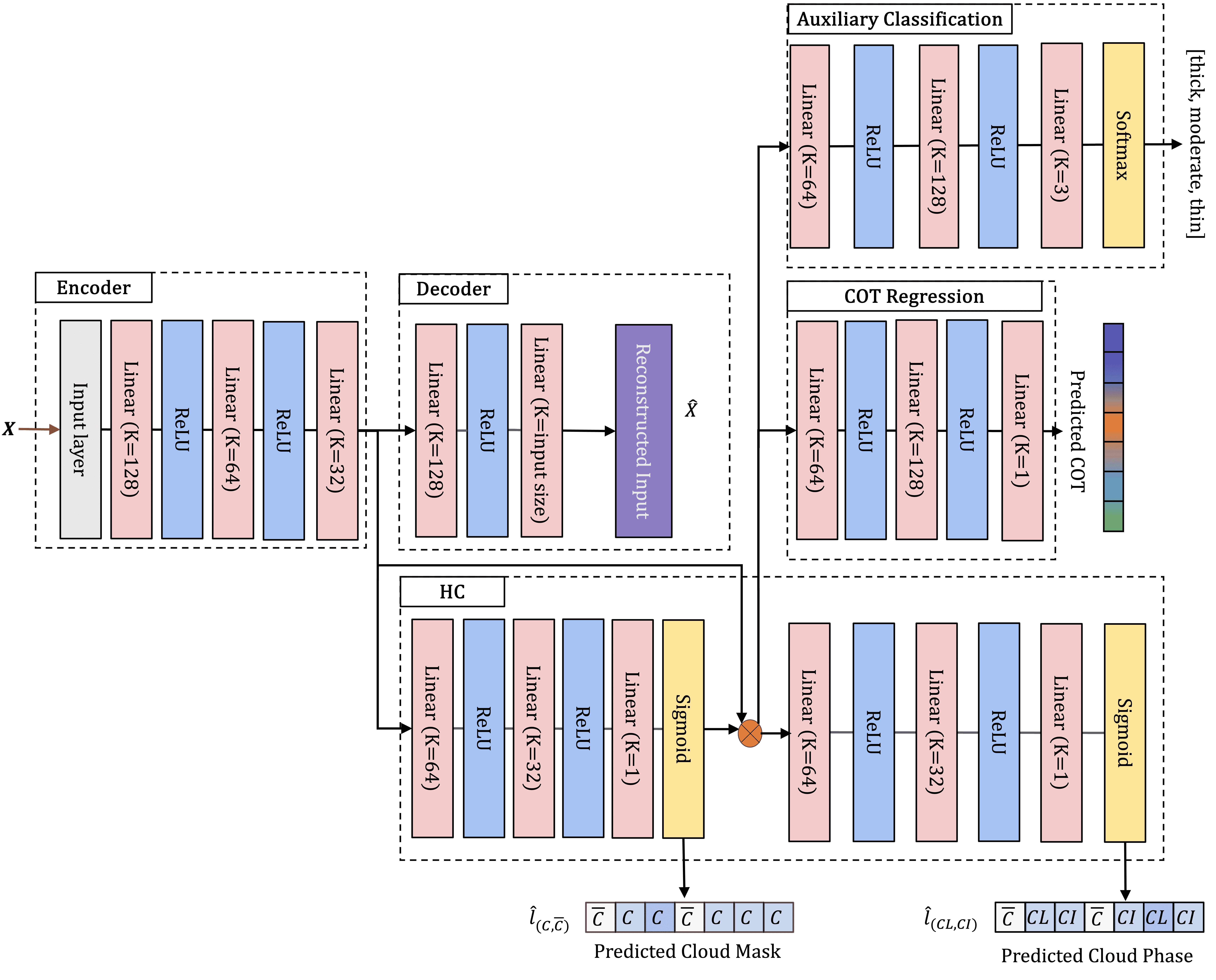}
\caption{The model MT-HCCR, where HC represents the proposed HC sub-network, CR represents the CAR model without the cross-attention module (A).}
\label{fig:mthccr}
\end{figure}

To provide a better understanding of the differences in models compared to the ablation study, below are figures of model architectures of each model. Figure~\ref{fig:seq} to Figure~\ref{fig:mthccr} illustrate model structures of \textbf{SEQ}, \textbf{MT-CR}, \textbf{MT-HCCR}, and \textbf{MT-HCCR}, respectively.

\section{Cross-Validation and Model Selection}
\label{sec:modelselection}

Based on the evaluation metrics and experiment results expounded in Section 5 in the main paper, our analysis extends to the application of the MT models to three different sensors simulated datasets, namely OCI, VIIRS, and ABI. Given the diverse evaluation metrics and the utilization of different satellite datasets, it is imperative to identify the singular optimal model among our multi-task learning-based models, namely MT-CR, MT-HCR, and MT-HCCAR, to be implemented on different datasets.

The process of model selection is based on the four metrics: ACC$_{bi}$, $AU(\overline{PRC})_w$, MSE, and $R^2$. The process includes 3 steps: 1) employing K-fold cross-validation, 2) executing one standard error rule (1SE rule), and 3) computing performance scores. Initially, K-fold cross-validation is employed to compare the model's absolute performance using the mean value of each metric over K folds. This involves aggregating the mean values of all four metrics across models trained on each dataset over K folds. Subsequently, the one standard error rule is applied to assess the balance between model absolute performance and computational complexity.

While the first and second steps prove effective in selecting the optimal model within the confines of a single metric and a single dataset, extending this evaluation to encompass all datasets and all metrics requires the introduction of absolute performance scores ($P_{ab}$ and 1SE performance scores), denoted as $P_{1SE}$. These proposed scores serve as comprehensive indicators for selecting a single model that excels across diverse datasets and evaluation metrics. Table~\ref{tab:model-selection} shows collected metrics and performance scores for each model and each dataset. The following paragraphs will describe K-fold cross-validation, one standard error rule, and our proposed performance score, respectively.

We note our model selection method is also useful for other multi-task learning applications due to the multi-objective nature of such applications and our method's capability of considering multiple metrics and datasets.

\begin{table*}[!ht]
    \centering
     \caption{Model selection using K-fold cross-validation and 1SE rule.}
     \label{tab:model-selection}
     \begin{tabular}{cccccrr}
     \hline
         \textsc{Dataset}	&	\textsc{Model}	&	$\mu(Metric)$	&	$SE(Metric)$	&		\textsc{1SE Range}  & $P_{ab}^{\mathcal{M,D}}$ & $P_{1SE}^{\mathcal{M,D}}$ \\
         \hline
         \hline
         & \multicolumn{5}{c}{\textsc{Mean value of ACC$_{bi}$ over 10 folds}}\\
     \hline
           \multirow{3}{*}{\textsc{OCI}} 
&	MT-CR	&	0.969	&	7.608E-04	&	[0.969,	0.970]		&	-1.54\%	&	0	\\																		
&	MT-HCR	&	0.984	&	\textbf{6.741E-04}	&	[0.983,	0.985]	&	 		-0.04\%	&	\textbf{1}	\\																	
&	MT-HCCAR	&	\textbf{0.985}	&	1.057E-03	&	[0.983,	0.986]	&		\textbf{0.00\%}	&	0	\\		
\hline
            \multirow{3}{*}{\textsc{VIIRS}}	
&	MT-CR	&	0.976	&	6.866E-04	&	[0.975,	0.976]	&	-1.39\%	&	0	\\																		
&	MT-HCR	&	0.989	&	5.027E-04	&	[0.988,	0.989]	&	-0.04\%	&	\textbf{1}	\\																		
&	MT-HCCAR	&	\textbf{0.989}	&	\textbf{4.961E-04}	&	[0.989,	0.990]	&		\textbf{0.00\%}	&	0	\\		\hline	

            \multirow{3}{*}{\textsc{ABI}}	
&	MT-CR	&	0.973	&	1.229E-03	&	[0.972,	0.974]	&		-1.37\%	&	0	\\																		
&	MT-HCR	&	0.986	&	\textbf{4.716E-04}	&	[0.986,	0.987]		&	-0.03\%	&	\textbf{1}	\\																	
&	MT-HCCAR	&	\textbf{0.987}	&	6.180E-04	&	[0.986,	0.987]	&	\textbf{0.00\%}	&	0	\\																		
         \hline
         \hline
         
& \multicolumn{5}{c}{\textsc{Mean value of $AU(\overline{PRC})_w$ over 10 folds}}\\
     \hline
           \multirow{3}{*}{\textsc{OCI}} 
&	MT-CR	&	0.953	&	1.60E-03	&	[0.951,	0.954]	&	-4.29\%	&	0	\\																	
&	MT-HCR	&	0.995	&	2.83E-04	&	[0.995,	0.995]	&	-0.04\%	&	0	\\																		
&	MT-HCCAR	&	\textbf{0.996}	&	\textbf{2.08E-04}	&	[0.995,	0.996]	&		\textbf{0.00\%}	&	\textbf{1}	\\		\hline	

            \multirow{3}{*}{\textsc{VIIRS}}	
&	MT-CR	&	0.939	&	3.17E-02	&	[0.908,	0.971]	&		-5.78\%	&	0	\\																
&	MT-HCR	&	0.996	&	1.88E-04	&	[0.996,	0.997]	&	-0.06\%	&	0	\\																		
&	MT-HCCAR	&	\textbf{0.997}	&	\textbf{1.37E-04}	&	[0.997,	0.997]	&		\textbf{0.00\%}	&	\textbf{1}	\\		\hline	

            \multirow{3}{*}{\textsc{ABI}}	
&	MT-CR	&	0.969	&	2.19E-03	&	[0.967,	0.971]	&	-2.75\%	&	0	\\																		
&	MT-HCR	&	\textbf{0.996}	&	\textbf{1.62E-04}	&	[0.996,	0.996]	&			\textbf{0.00\%}	&	\textbf{1}	\\																		
&	MT-HCCAR	&	\textbf{0.996}	&	1.67E-04	&	[0.996,	0.996]	& -0.02\%	&	0	\\	

\hline
\hline

& \multicolumn{5}{c}{\textsc{Mean value of MSE over 10 folds}}\\
     \hline
           \multirow{3}{*}{\textsc{OCI}} 
&	MT-CR	&	0.055	&	2.916E-03	&	[0.058,	0.052]	&		-100.00\%	&	0	\\																		
&	MT-HCR	&	0.034	&	5.970E-04	&	[0.034,	0.035]	&	-25.27\%	&	0	\\																		
&	MT-HCCAR	&	\textbf{0.027}	&	\textbf{4.910E-04}	&	[0.028,	0.027]	&		\textbf{0.00\%}	&	\textbf{1}	\\		\hline		

            \multirow{3}{*}{\textsc{VIIRS}}	
&	MT-CR	&	0.043	&	2.075E-03	&	[0.045,	0.041]	&	-76.48\%	&	0	\\																		
&	MT-HCR	&	0.030	&	\textbf{8.104E-04}	&	[0.031,	0.029]	&	 	-21.65\%	&	0	\\																	
&	MT-HCCAR	&	\textbf{0.025}	&	1.282E-03	&	[0.026,	0.023]	&			\textbf{0.00\%}	&	\textbf{1}	\\		
\hline	

            \multirow{3}{*}{\textsc{ABI}}	
&	MT-CR	&	0.057	&	1.519E-03	&	[0.059,	0.056]		&	-78.86\%	&	0	\\																		
&	MT-HCR	&	0.038	&	5.714E-04	&	[0.039,	0.038]		&	-19.48\%	&	0	\\																		
&	MT-HCCAR	&	\textbf{0.032}	&	\textbf{3.121E-04}	&\textbf{	[0.032,	0.032]}	&		\textbf{0.00\%}	&	\textbf{1}	\\		

\hline
\hline

& \multicolumn{5}{c}{\textsc{Mean value of $R^2$ over 10 folds}}\\
     \hline
           \multirow{3}{*}{\textsc{OCI}} 
&	MT-CR	&	0.758	&	1.24E-02	&	[0.770,	0.746]		&	-14.25\%	&	0	\\																	
&	MT-HCR	&	0.854	&	2.55E-03	&	[0.857,	0.851]		&	-3.39\%	&	0	\\																		
&	MT-HCCAR	&	\textbf{0.884}	&	\textbf{2.35E-03}	&	[0.882,	0.886]	&	\textbf{0.00\%}	&	\textbf{1}	\\		\hline		

            \multirow{3}{*}{\textsc{VIIRS}}	
&	MT-CR	&	0.810	&	9.62E-03	&	[0.800,	0.820]		&	-9.72\%	&	0	\\																		
&	MT-HCR	&	0.874	&	\textbf{3.59E-03}	&	[0.870,	0.878]	&	 	-2.58\%	&	0	\\																		
&	MT-HCCAR	&	\textbf{0.897}	&	5.30E-03	&	[0.892,	0.902]	&			\textbf{0.00\%	}&	\textbf{1}	\\		\hline														

            \multirow{3}{*}{\textsc{ABI}}	
&	MT-CR	&	0.751	&	7.44E-03	&	[0.743,	0.758]		&	-13.25\%	&	0	\\																		
&	MT-HCR	&	0.840	&	2.20E-03	&	[0.838,	0.843]	&		-2.91\%	&	0	\\																		
&	MT-HCCAR	&	\textbf{0.866}	&\textbf{	1.51E-03}	&	[0.864,	0.867]		&	\textbf{0.00\%	}&	\textbf{1}	\\																		
     \hline
     \end{tabular}
     \label{tab:kfold_reg}
 \end{table*}

\subsection{K-fold cross-validation and 1SE rule}
\textbf{K-fold cross-validation}
is used to compare the absolute performance of each model $\mathcal{M}$ trained by each dataset $\mathcal{D}$ based on mean values of the four evaluation metrics over K folds. We use the number of folds $K=10$ because this value has been found through experiments to generally result in a model skill estimate with low bias and modest variance \cite{kuhn2013applied,james2013introduction}. For each fold, $(K-1)N/K$ pixels are in the training set, and the left $N/K$ pixels are in the test set. \textbf{Mean values} $\mu(Metric)$ of four evaluation metrics over 10 folds are calculated for each experiment to compare differences in model absolute performance. Among the four metrics, ACC$_{bi}$ and $AU(\overline{PRC})_w$ access the classification task while MSE and $R^2$ access the regression task. The mean value of a certain metric $Metric$ over 10 folds $\mu(Metric)$ is: 
$\mu(Metric) = \frac{1}{K}\sum_{k=1}^{K=10}Metric^{k, \mathcal{M}^{\mathcal{D}}}$, where $Metric^{k}$ is $Metric$ for fold $k$ of model $\mathcal{M}$ trained by dataset $\mathcal{D}$. 

With the mean values from K-fold cross-validation, we then use \textbf{one standard error rule (1SE rule)} \cite{krstajic2014cross} to select the most parsimonious model for each dataset. The mean of evaluation metrics, computed over 10 folds, provides an absolute measure of model accuracy. However, it is essential to consider the trade-off between model complexity and absolute performance, as this balance is crucial for assessing the overall quality of a model. Therefore, in addition to the mean value, we also calculate the region defined by one standard error rule (1SE region) of each metric. With the 1SE range of each metric, we need to select the ``1SE region of the best'', which is the 1SE range of the model with minimum mean error. Then the simplest model whose mean falls within the 1SE region of the best is the most parsimonious model according to the 1SE rule \cite{friedman2010regularization,russell2010artificial}. Under our experiment setting, the \textbf{standard error} for a certain metric $SE(Metric)$ is calculated as: 
\begin{equation}
    SE(Metric) = \frac{s}{\sqrt{K}},
\end{equation}
\noindent where $s$ is the square root of variance $s^2$ for $Metric$ over 10 folds. Then the 1SE region of each model is defined as: $[\mu-SE, \mu+SE]$. 

In our experimental analyses, we derive the following conclusion regarding the application of the 1SE rule: if none of the $\mu(Metric)$ values for alternative models lie within the 1SE region of the optimal model, the model exhibiting the highest mean metric value is the preferred choice based on this specific metric.

\subsection{The absolute score and the 1SE rule score} 

Despite conducting K-fold cross-validation and 1SE rule, we still need a method to describe the general model performance across all four evaluation metrics and all three datasets. Here we introduced the absolute performance score and the 1SE rule score.

To describe the absolute performance based on the mean value as well as the model selection results based on the 1SE rule for each model over all datasets and all metrics, we propose to use \textbf{absolute performance score $P_{ab}$} and \textbf{1SE score $P_{1SE}$}. The absolute performance score $P_{ab}$ describes how worse other models are than the model with the best absolute performance. The performance score for each model is calculated by the following formula: 
\begin{equation}
\label{eq:absolutescore}
    \begin{split}
    P_{ab}^{\mathcal{M}} &=  \sum_{D}\sum_{Metric} P_{ab}^{\mathcal{M,D}}\\
&= \sum_{D}\sum_{Metric} \frac{\mu({Metric})^{\mathcal{M}^{\mathcal{D}}} - \mu({Metric})^{\mathcal{M}^{\mathcal{D}\star}_{\mu}}}{\mu({Metric})^{\mathcal{M}^{\mathcal{D}\star}_{\mu}}},
    \end{split}
\end{equation}

\noindent where components $P_{ab}^{\mathcal{M, D}}$ is the score for a specific model $\mathcal{M}$ applied to a specific dataset $\mathcal{D}$, $\mu({Metric})^{\mathcal{M}^{\mathcal{D}}}$ is the mean value of a specific metric $Metric$ of model $\mathcal{M}^{\mathcal{D}}$ applied to dataset $\mathcal{D}$, and $\mathcal{M}^{\mathcal{D}\star}_{\mu}$ is the model with the best mean value for dataset $\mathcal{D}$.

The 1SE performance score is the weighted sum of counts that the model is selected according to the 1SE rule across all metrics and all datasets. The component $P_{1SE}^{\mathcal{M,D}}$ is the 1SE performance score for model $\mathcal{M}$ applied to dataset $\mathcal{D}$. The calculation of the total score $P_{1SE}$ for model $M$ is: 
\begin{equation}
\label{eq:1sescore}
    \begin{split}
    P_{1SE}^{\mathcal{M}} &= \sum_{D}\mathbf{w_m}\sum_{Metric} P_{1SE}^{\mathcal{M,D}} \\
    & = \sum_{D}\sum_{Metric} \psi(\mu(Metric), \mu(\mathcal{M}, \mathcal{D} ),
    \end{split}
\end{equation}
\noindent where $\psi(\cdot)$ is a rule that if Model $\mathcal{M}$ trained by dataset $\mathcal{D}$ is selected by 1SE rule based on the mean value of a certain metric, then $\psi(\cdot)$ outputs 1, and otherwise 0. Weights $\mathbf{w_m} = [w_{Acc_{bi}}, w_{AU(\overline{PRC})_w}, w_{MSE}, w_{R^{2}}] $ define the significance of four metrics. We set $\mathbf{w_m} = [1,1,1,1]$ in our experiments because we consider the classification task and regression task equally important in our application. 
We will further discuss the model performance shown in Table ~\ref{tab:model-selection} based on 1SE region of the best to show that MT-HCCAR is optimal based on all four metrics of ACC$_{bi}$, $AU(\overline{PRC})_w$, MSE, and $R^2$.

\subsection{Model selection results}
\label{sec:selectionresults}

This section discusses the superior performance of our proposed MT-HCCAR model, evident in its highest absolute performance score $P_{ab}^{\mathcal{M}}$ score and the top 1SE rule score $P_{1SE}^{\mathcal{M}}$. The last two columns of Table~\ref{tab:model-selection} are components of $P_{ab}^{\mathcal{M}}$ and $P_{1SE}^{\mathcal{M}}$. Utilizing Equation~\ref{eq:absolutescore}, the absolute performance scores for each model are calculated as follows: $P_{ab}^{{MT-CR}} = -4.12\%$, $P_{ab}^{{MT-HCR}} = -0.21\%$, and $P_{ab}^{{MT-HCCAR}} = -0.02\%$. The highest $P_{ab}^{\mathcal{M}}$ indicates that MT-HCCAR exhibits the best mean value across all evaluation metrics, including ACC$_{bi}$, $AU(\overline{PRC})_w$, and $R^2$, across all satellite datasets. 

Furthermore, MT-HCCAR achieves the most substantial $P_{1SE}^{\mathcal{M}}$. By using Equation~\ref{eq:1sescore}, the 1SE performance scores for each model are calculated as follows: $P_{1SE}^{{MT-CR}} = 0$, $P_{1SE}^{{MT-HCR}} = 4$, $P_{1SE}^{{MT-HCCAR}} = 8$. Achieving the highest $P_{1SE}^{\mathcal{M}}$ suggests that all other models exhibit mean values beyond the 1 standard error (1SE) range of MT-HCCAR. This implies that MT-HCCAR avoids overfitting in comparison to other models, considering model complexity. This finding further establishes MT-HCCAR as the optimal model for implementation across various satellite datasets.

Additionally, MT-HCCAR exhibits higher mean values and a narrower 1SE range compared to alternative models, indicating superior generalization. In summary, MT-HCCAR, adhering to the 1SE rule, demonstrates superior performance in both classification and regression tasks, surpassing other models without succumbing to overfitting.

\end{document}